# Performance of Stanford and Minipar Parser on Biomedical Texts

Rushdi Shams
*Department of Computer Science, University of Western Ontario, London, ON N6A 5B7, Canada*
*rshams@uwo.ca*

## Abstract

*In this paper, the performance of two dependency parsers, namely Stanford and Minipar, on biomedical texts has been reported. The performance of the parsers to assign dependencies between two biomedical concepts that are already proved to be connected is not satisfying. Both Stanford and Minipar, being statistical parsers, fail to assign dependency relation between two connected concepts if they are distant by at least one clause. Minipar's performance, in terms of precision, recall and the F-Score of the attachment score (e.g., correctly identified head in a dependency), to parse biomedical text is also measured taking the Stanford's as a gold standard. The results suggest that Minipar is not suitable yet to parse biomedical texts. In addition, a qualitative investigation reveals that the difference between working principles of the parsers also play a vital role for Minipar's degraded performance.*

## I. Introduction

Dependency parsers, unlike constituency parsers, follow the dependency grammar that is inspired by a basic assumption that the syntactic structure comprises words linked by some binary and asymmetrical relations called dependency relations. Most of the dependency parsers express the relation between two words in the form of a triplet like *dependency (head, dependent)*, where *dependency* is the relation and *head* is the word that has been modified by the *dependent.* Dependency parsers, now-a-days, are popular in natural language processing for a number of reasons: as they provide an implicit predicate-argument structure of the sentences, they play an important role in machine translation and information extraction; people with limited knowledge on linguistics can achieve a deeper understanding on the usage of language and its development; and last but not the least, dependency parsers lead to the development of effective syntactic parsers for a number of domains [1] like biomedical and bioinformatics. The extraction of connected biomedical concepts (i.e., disease, treatment, genes) from texts has drawn the attention of the scientists interested in finding functional similarity (i.e., identification of genes involved in human diseases) [2]. To achieve this, researchers are currently not only using dependency parsers for their ability to extract the links among the words but also developing dependency grammar based corpora like BioInfer [3].

Although the underlying theory of the dependency parsers is the same, their working principles vary for a number of reasons: some dependency parsers, like Stanford Parser, modified the original grammar rules to introduce semantics [4]; several parsers, like Stanford, Minipar and Link parsers, use different techniques to find the heads of a sentence [5]; the size and the domain of the training corpus of the parsers vary- like Stanford is trained on the large Penn Wall Street Journal Corpus and Susanne, a wide coverage-small sized corpus was used to train Minipar [6]. Such different working principles have subtle impact on the performances of these dependency parsers. For instance, Comelles *et al.* [7] reported a comparative and a qualitative analysis on five popular dependency parsers where they found several linguistic errors produced by the parsers. Besides, although Katrenko and Adriaans [8] sustained several drawbacks of the use of the dependency parsers in a specific domain (e.g., parsing biomedical text), domain-specific use of the parsers has been reported in number of research [9-12].

This paper reports a quantitative and a qualitative analysis on two dependency parsers, namely Stanford and Minipar, to evaluate their performance in parsing biomedical text. Given a set of 40 pairs of connected concepts from four biomedical texts [13], the parsers are tested to see if they are able to find out the concepts as connected. The comparison reports that Stanford parser performed better than Minipar in finding connected concepts. For every pair of connected concepts in this set, the sentences of the text that contain both of them are fed to the parsers to get the dependencies in a triplet form like *dependency (head, dependent)*. Taking this output of the Stanford Parser as a gold standard, the attachment score of Minipar parser, which is the percentage of words that have the correct head, is measured. Then, the F-score (e.g., the equally weighted harmonic mean of the precision and recall) of the attachment scores is calculated. The results show that Minipar performed consistently for the whole set of the connected concepts but its performance on biomedical text is not satisfactory.

In the remainder of the paper, brief introductions to dependency grammar and the chosen parsers are provided in Section II. Section III describes the performance of the parsers on the biomedical texts. Finally, Section IV concludes the paper.

## II. Background

In this section, a brief description of dependency grammar and the working principles of the parsers considered for this experimentation, namely Stanford and Minipar, are presented.

### *A. Dependency Grammar*

Since the commencement of the idea to present syntactic structure of a sentence by linking words with a number of dependencies, many variations are proposed. But the basic assumption of dependency grammar remains unchanged: it is possible to relate the words in a sentence by a number of asymmetric and binary dependency relations. Moreover, the words have very specific roles in a dependency relation with one another: the constituent word of a dependency relation can be either a head or a dependent that modifies the head. If we take the sentence *Economic news had little effects on financial markets* as an example, then the dependency relations can be derived from the Direct Acyclic Graph (DAG) drawn from the sentence. The DAG drawn for every sentence has some properties: the words in the sentence are the nodes of the graph and their relations are asymmetric edges that connect them. However, according to the theory of dependency grammar, a word can modify more than one word but can be modified by at most one word. So, in the DAG, a word can have many outgoing edges to the words that it is modifying but can have only one incoming edge from the word that modifies it. Fig. 1 shows the DAG drawn for the sentence in our example [1].

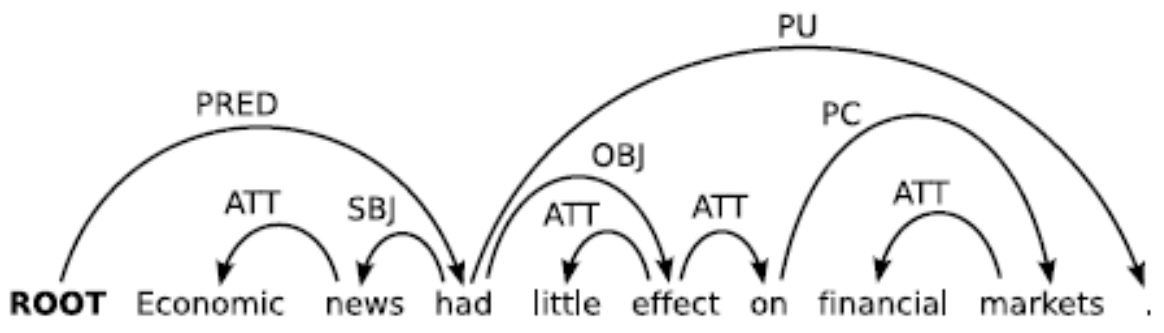

Figure 1. Direct Acyclic Graph (DAG) for a given English Sentence

From Fig. 1, we see that the DAG contains a ROOT, known as the head of the sentence in the dependency grammar, is a mandatory node in the tree and does not modify any word but can be modified by others. In Fig. 1, we account many dependency relations which are the labels of the arcs. For example, *news* is the *dependent* that modifies the *head* called *economic* by a dependency relation *ATT* (shorthand for *attribute*). In dependency grammar, such a relation is written in the form of a triplet: *dependency (head, dependent)* or *dependency (dependent, head)*. Therefore, the dependency between *economic* and *news* will be expressed as *ATT (economic, news)* or *ATT (news, economic).* The corresponding phrase structure tree, used by most of the constituency parsers, of the given sentence can be seen in Fig. 2 [1].

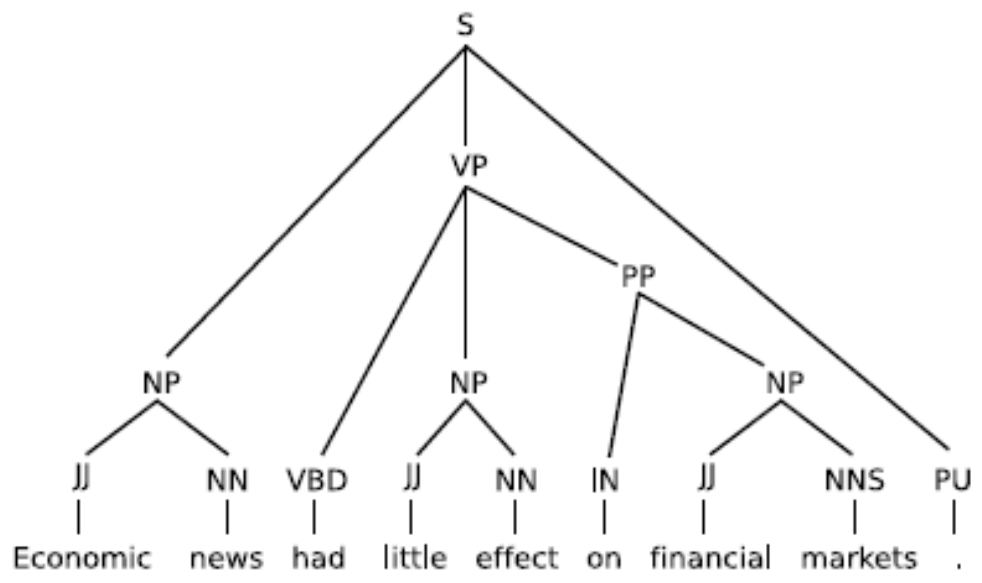

Figure 2. Phrase Structure Tree for a given English Sentence

The difference between two representations is obvious: dependency grammar represents the head-dependent relations among the words by classifying them with *functional* categories like subject and object while the phrase structure grammar represents the relations among constituents or phrases by classifying them into *structural* categories like noun phrase (NP) or verb phrase (VP).

### *B. The Stanford and Minipar Parsers*

Several parsers have been developed to represent the dependency relations in a sentence. Minipar, a statistical parser, descended from its previous installment called Principar [6] and it works based on the basic principles of dependency grammar. It was developed in 1998 and immediately went under strict evaluation on the Susanne Corpus. The evaluation outcome was satisfying as it extracted 79 percent of the dependency relations in the corpus with a high precision of 89 percent. The key advantages of Minipar are: it uses the basic principles of dependency grammar without any modifications, its availability, its simplicity, its performance and its training corpus which was a subset of Brown Corpus that makes it a wide-coverage parser [6]. Minipar drew the attention of the researchers as soon as its performance is published and till to date, its use is manifolds: to parse domain specific texts, to compete in natural language parsers' evaluations and to be counted as a gold standard in many other parser evaluations.

The developers of the Stanford parser originally developed it in 2003 as a statistical constituency parser but its working principle significantly changed when Stanford Dependency (SD) scheme was developed in 2006. The idea of creating a scheme like SD, which is a modified modern version of the early dependency grammar, was revolutionary and the parser started to produce more significant and meaningful dependency relations. Several evaluations on SD also reveal that the scheme not only brings appropriate syntactic dependency relations but also is capable of relating words semantically. The SD scheme is now reported as a widely used grammar scheme for parsing the texts of the domains like biomedical and bioinformatics.

A brief summary of the Stanford and Minipar parsers is provided in Table 1.

| Parser Name | Version Used | Source Language | Training on |
|---|---|---|---|
| Stanford Parser | 1.6.5 | Java | Penn Wall Street Journal |
| Minipar Parser | 0.5 | C++ | Susanne Corpus |

Table 1. Summary of Stanford and Minipar Parser

## III. Performance of the Parsers on Biomedical Texts

### *A. Setup*

To test the parsers, the set of connected pairs of biomedical concepts from four scholarly articles has been used. For every pair in the set, the sentences of the papers that contain both of these concepts will be fed to the parsers. However, the parsers are not trained with biomedical corpora. That is why if we use the Part of Speech (POS) taggers of these parsers to tag biomedical text, then it is likely that they might assign wrong POS assignments. Fig. 3, for example, shows that *glutamate* is said to be an *adjective (JJ)* by the Stanford Tagger though it is a chemical component and a *noun (NN)* tag will be appropriate.

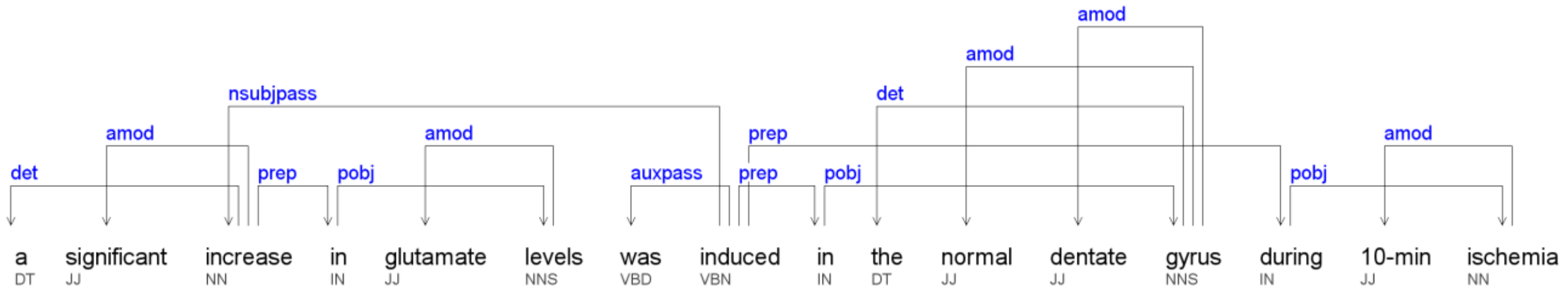

Figure 3.Inappropriate POS Tagging by the Stanford Tagger

Therefore, before feeding the sentences to the parsers, the sentences are tagged with Genia POS Tagger [14], which is trained with biomedical corpora and designed specifically for tagging biomedical text. The working procedure that is followed during this experiment is shown in Fig. 4.

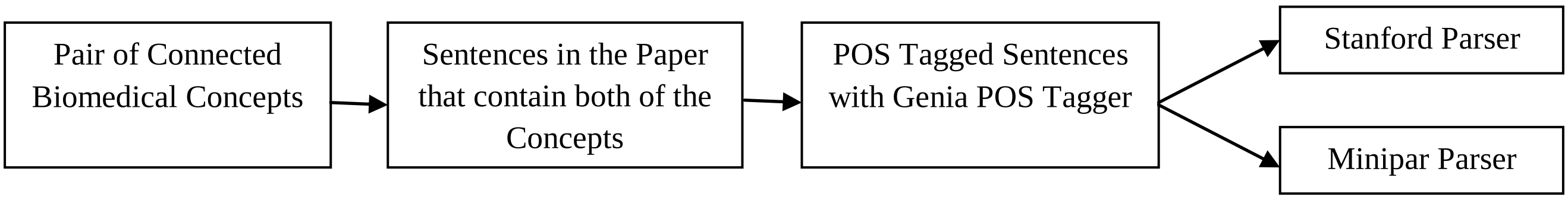

Figure 4. Working Methodology

### *B. Quantitative Analysis*

First, for every pair of concepts in the given set, the sentences containing both of the concepts are fed to the parsers to assess the ability of them to find out the pairs of concepts as connected. For example, for the connected concepts *Ischemia* and *Glutamate*, the sentences containing both of them are fed to the parsers and individual record of the parsers has been kept when the parsers find a triplet like *dependency (Ischemia, Glutamate)* or *dependency (Glutamate, Ischemia)*.

The detailed record on the parsers' ability to find the given concepts as connected with a dependency relation is shown in Table 2.

| Connection | Total Sentences | Stanford | Minipar |
|---|---|---|---|
| Ischemia-Glutamate | 23 | 2 | 0 |
| Levels-Ischemia | 15 | 2 | 0 |
| Levels-Glutamate | 16 | 11 | 8 |
| Glutamate-Neurons | 9 | 0 | 0 |
| 10Min-Ischemia | 17 | 10 | 7 |
| CA4-Glutamate | 8 | 0 | 0 |
| Increase-Glutamate | 11 | 0 | 0 |
| 10Min-Glutamate | 13 | 0 | 0 |
| 5Min-Ischemia | 6 | 5 | 3 |
| 5Min-Glutamate | 5 | 0 | 0 |

(a)

| Connection | Total Sentences | Stanford | Minipar |
|---|---|---|---|
| Friedreich-Ataxia | 11 | 8 | 6 |
| PDHC-Ataxia | 9 | 1 | 0 |
| Activity-Friedreich | 7 | 0 | 0 |
| Patients-Ataxia | 7 | 3 | 1 |
| Activity-Ataxia | 7 | 1 | 0 |
| PDHC-Friedreich | 7 | 1 | 0 |
| Preparations-Ataxia | 4 | 0 | 0 |
| Preparations-Friedreich | 4 | 0 | 0 |
| Pyruvate-Ataxia | 5 | 1 | 0 |
| Patients-Friedreich | 6 | 1 | 0 |

(b)

| Connection | Total Sentences | Stanford | Minipar |
|---|---|---|---|
| AAS-Treatment | 11 | 1 | 0 |
| Use-AAS | 17 | 9 | 2 |
| AAS-Testosterone | 8 | 2 | 1 |
| Gonadotropin-Treatment | 9 | 0 | 0 |
| Testosterone-Treatment | 10 | 4 | 3 |
| Levels-Testosterone | 12 | 7 | 2 |
| AAS-Conditions | 5 | 0 | 0 |
| Treatment-HCG | 5 | 1 | 0 |
| Replacement-Therapy | 5 | 3 | 3 |
| Treatment-Therapy | 5 | 2 | 2 |

(c)

| Connection | Total Sentences | Stanford | Minipar |
|---|---|---|---|
| Inhibition-GABA | 33 | 23 | 11 |
| GABA-Synapse | 20 | 1 | 0 |
| Neurons-Synapse | 9 | 0 | 0 |
| Inhibition-Hippocampus | 6 | 1 | 0 |
| Synapse-Change | 5 | 0 | 0 |
| Neurons-GABA | 16 | 3 | 1 |
| Properties-GABA | 3 | 0 | 0 |
| GABA-Change | 3 | 0 | 0 |
| GABA-Number | 4 | 2 | 1 |
| Cl-Gradient | 3 | 3 | 3 |

(d)

Table 2. The Ability of Stanford and Minipar to find the Dependencies among Connected Concepts from Paper on (a) Ischemia and Glutamate (b) Ataxia and Dehydrogenase (c) Hypogonadism and Gonadotropin (d) Epilepsy and GABA

From Table 2, we can see that the parsers found dependencies between the connected concepts more often if the distance between the given concepts is short, to be more exact, if they are in the same clause. For example, *Glutamate-Levels, Friedreich-Ataxia, Use-AAS* and *Inhibition-GABA* are the four pairs of concepts that the parsers found to be connected most of the times. From careful observations, we can see that the pairs of concepts maintain a very short syntactic distance with each other: *Glutamate* mostly acts as a noun compound modifier of *Levels* (e.g., *the Glutamate levels increased during the 10 minute Ischemia*), *Friedreich* mostly is a lexical modifier of *Ataxia* (e.g., *the Friedreich's Ataxia is a kind of brain disease*), *AAS* is a steroid and is mostly modified by *Use* (e.g., *the use of AAS can affect the release of testosterone*), and *GABA* mostly acts as a noun compound modifier of *Inhibition* (e.g., *the GABA inhibition was manifested during the observation*). From this observation, we can come to a decision that statistical parsers struggle to find the relations between two words if they are at least one clause away from each other.

The previous work [13] assumed that it would find the concepts that hold explicit or implicit semantic connections but it came up with some pairs of concepts that hardly hold such semantic connections. However, the pairs of concepts in Table 2 to which the parsers could not assign any dependency relation have higher possibilities to hold the semantic relations. For example, the pairs *Ischemia-Glutamate, PDHC-Ataxia, AAS-Testosterone* and *Inhibition-Hippocampus* are semantically connected according to the UMLS Semantic Relations Networks [15] but the parsers have a low success in relating them with a dependency relation.

Second, the output of the Stanford Parser (henceforth, *Key*) has been considered as the gold standard and the output of Minipar Parser (henceforth, *Answer*) has been compared with the *Key* to find out the percentage of words that have the correct heads (e.g., Attachment Score). For example, if the *Answer* has the head *Ischemia*, so has the *Key*, then *Ischemia* is called correctly identified head in the *Answer* (e.g., Attachment score of *Answer* increases). As both of the parsers have different sets of dependency relations, only the head but both of the head and the relation, is considered to calculate the attachment score.

After calculating the attachment score of the *Answer* for any given pair of connected concepts, the precision and recall of the attachment score is calculated. The precision of the score is:

$$Precision = \frac{True\ Positives}{True\ Positives + False\ Positives} \quad (1)$$

Where, $True\ Positives$ are the number of heads present both in the *Answer* and the *Key* and $False\ Positives$ are the number of heads present in the *Answer* but in the *Key*.

The recall of the score is:

$$Recall = \frac{True\ Positives}{True\ Positives + False\ Negatives} \quad (2)$$

Where, $False\ Negatives$ are the number of heads present in the *Answer* but in the *Key*.

Finally, the F-Score of the *Answer*, which is the equally weighted harmonic mean of the precision and the recall, has been measured:

$$F - Score = 2 \times \frac{Precision \times Recall}{Precision + Recall} \times 100\% \quad (3)$$

For every pair of connected concepts in the set, Table 3 shows the precision, the recall, and the F-score.

| Connection | Precision | Recall | F-Score | Connection | Precision | Recall | F-Score |
|---|---|---|---|---|---|---|---|
| Ischemia-Glutamate | 37.92 | 19.24 | 25.52 | Friedreich-Ataxia | 34.23 | 20.76 | 25.84 |
| Levels-Ischemia | 38.46 | 17.77 | 24.30 | PDHC-Ataxia | 38.46 | 25.97 | 31.00 |
| Levels-Glutamate | 36.02 | 19.85 | 25.59 | Activity-Friedreich | 31.91 | 18.75 | 23.62 |
| Glutamate-Neurons | 41.17 | 12.98 | 19.73 | Patients-Ataxia | 37.57 | 24.48 | 29.64 |
| 10Min-Ischemia | 36.49 | 19.90 | 25.75 | Activity-Ataxia | 31.91 | 18.75 | 23.62 |
| CA4-Glutamate | 40.90 | 14.51 | 21.42 | PDHC-Friedreich | 34.84 | 19.40 | 24.92 |
| Increase-Glutamate | 46.87 | 14.63 | 22.29 | Preparations-Ataxia | 35.13 | 18.84 | 24.52 |
| 10Min-Glutamate | 39.65 | 22.63 | 28.81 | Preparations-Friedreich | 35.13 | 18.84 | 24.52 |
| 5Min-Ischemia | 61.78 | 21.46 | 31.85 | Pyruvate-Ataxia | 34.35 | 26.16 | 29.70 |
| 5Min-Glutamate | 36.28 | 36.84 | 36.55 | Patients-Friedreich | 36.08 | 18.42 | 24.38 |
| AAS-Treatment | 41.26 | 26.55 | 32.30 | Inhibition-GABA | 44.3 | 31.16 | 36.58 |
| Use-AAS | 40.06 | 23.36 | 29.51 | GABA-Synapse | 44.87 | 20.65 | 28.28 |
| AAS-Testosterone | 38.07 | 24.34 | 29.69 | Neurons-Synapse | 42.25 | 28.93 | 34.34 |
| Gonadotropin-Treatment | 51.21 | 33.22 | 40.29 | Inhibition-Hippocampus | 31.46 | 19.04 | 23.72 |
| Testosterone-Treatment | 39.86 | 19.40 | 26.09 | Synapse-Change | 42.40 | 37.50 | 39.79 |
| Levels-Testosterone | 35.14 | 19.08 | 24.73 | Neurons-GABA | 41.77 | 20.37 | 27.38 |
| AAS-Conditions | 38.97 | 29.77 | 33.75 | Properties-GABA | 43.33 | 22.67 | 29.76 |
| Treatment-HCG | 39.04 | 27.66 | 32.37 | GABA-Change | 37.42 | 34.65 | 35.98 |
| Replacement-Therapy | 53.04 | 38.85 | 44.84 | GABA-Number | 50.00 | 34.61 | 40.90 |
| Treatment-Therapy | 48.69 | 27.31 | 34.99 | Cl-Gradient | 25.00 | 35.08 | 29.19 |

Table 3: Precision, Recall, and F-Score of the Attachment Scores of Minipar

The F-Score of the attachment scores of Minipar shows that the parser maintains a consistent precision and recall throughout the papers. However, its low precision and recall in this case compared to the evaluation on Susanne corpus suggests that the parser is not ready yet to effectively parse biomedical texts.

## *C. Qualitative Analysis*

Both Stanford and Minipar parsers are trained on corpora in which not many questions occur [5] [6]. Although in biomedical texts not many interrogative sentences occur, but when occur, the output of the parsers differ largely that contributes to the low attachment score of Minipar. For example, the parser outputs for an interrogative sentence are shown in Fig. 5. The Stanford dependency trees are generated by a modified visualization tool offered by Athar [16] and the Minipar dependency trees are generated by a visualization tool offered by University of Zurich [17].

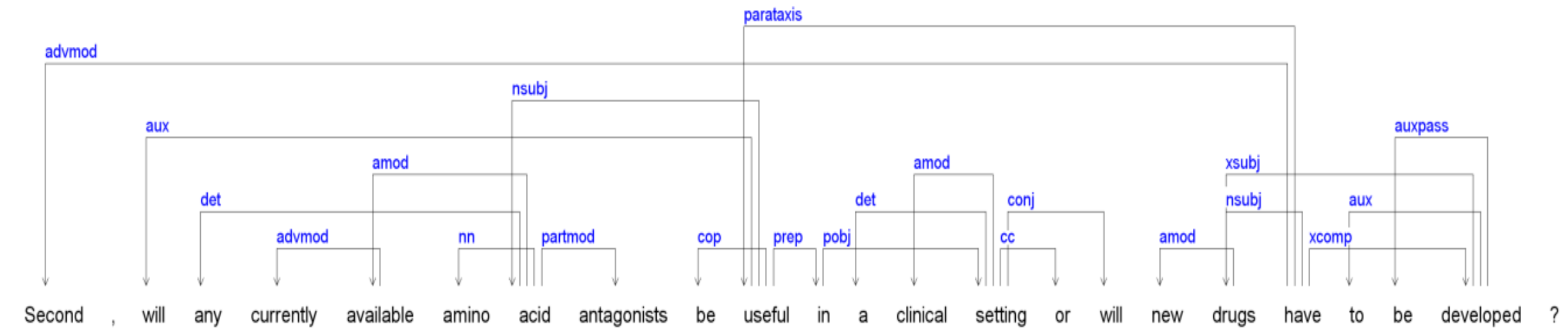

(a)

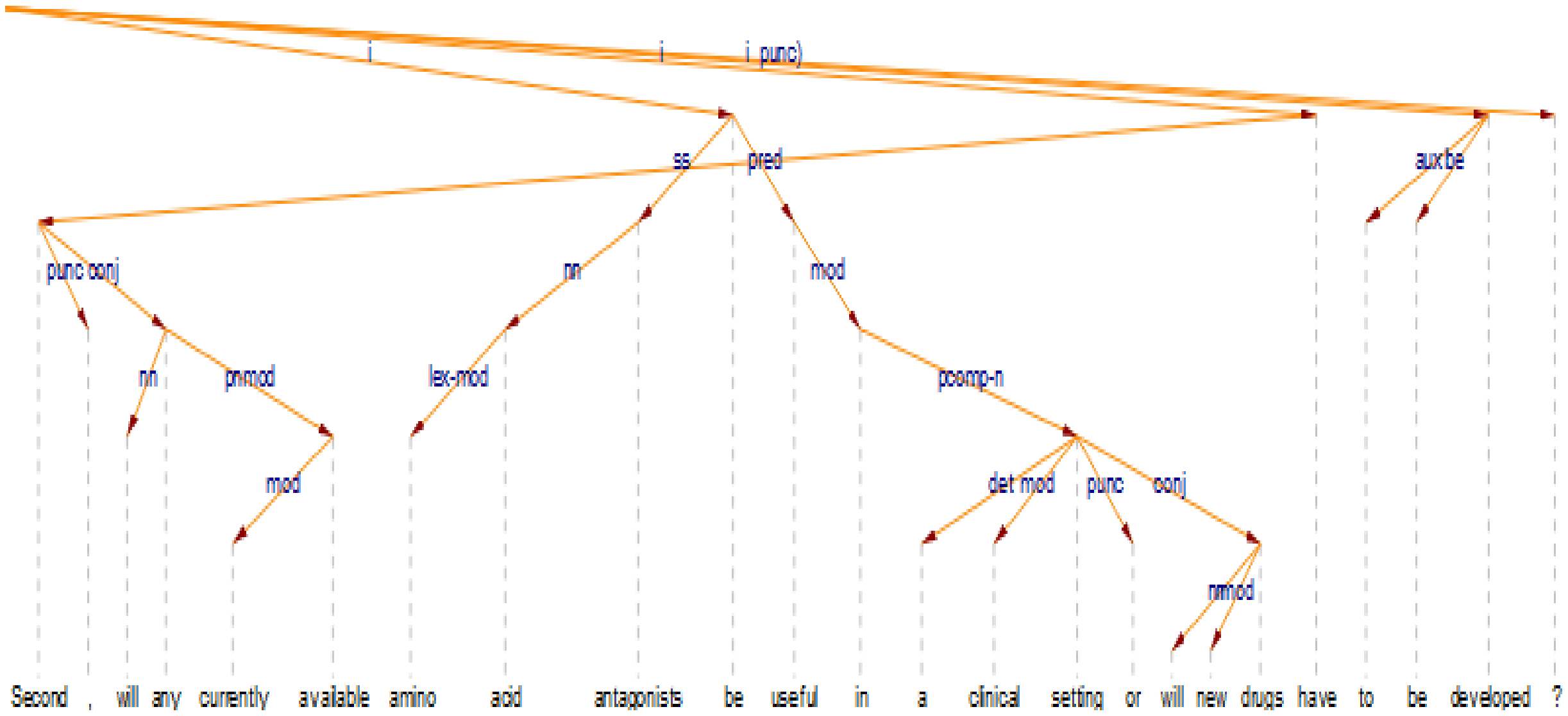

(b)

Figure 5. (a) Stanford and (b) Minipar generate Different Dependency Trees for Questions

Unlike Stanford, Minipar considers quotation marks while parsing but its dependency tree begins to differ with Stanford as it comes across such sentences. Fig. 6 shows how the dependency tree differs when the parsers parse sentences with quotation marks.

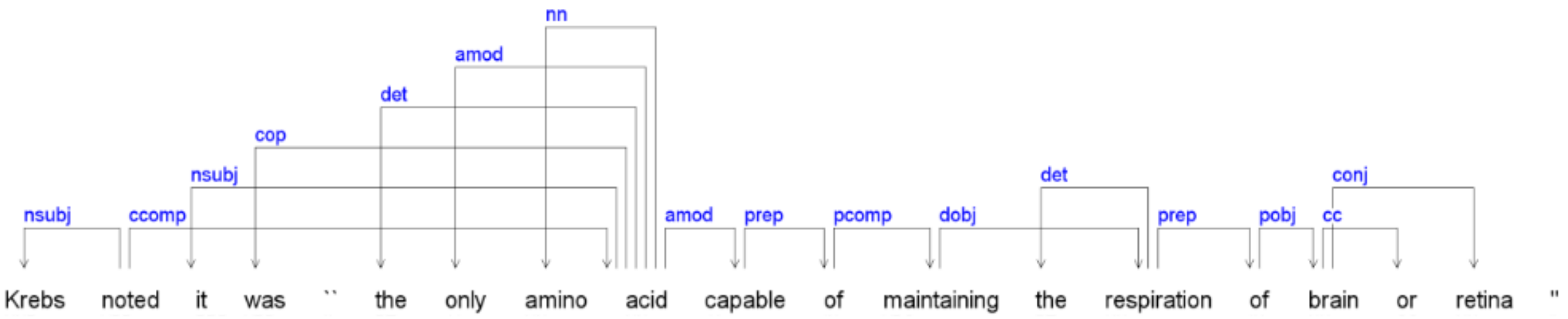

(a)

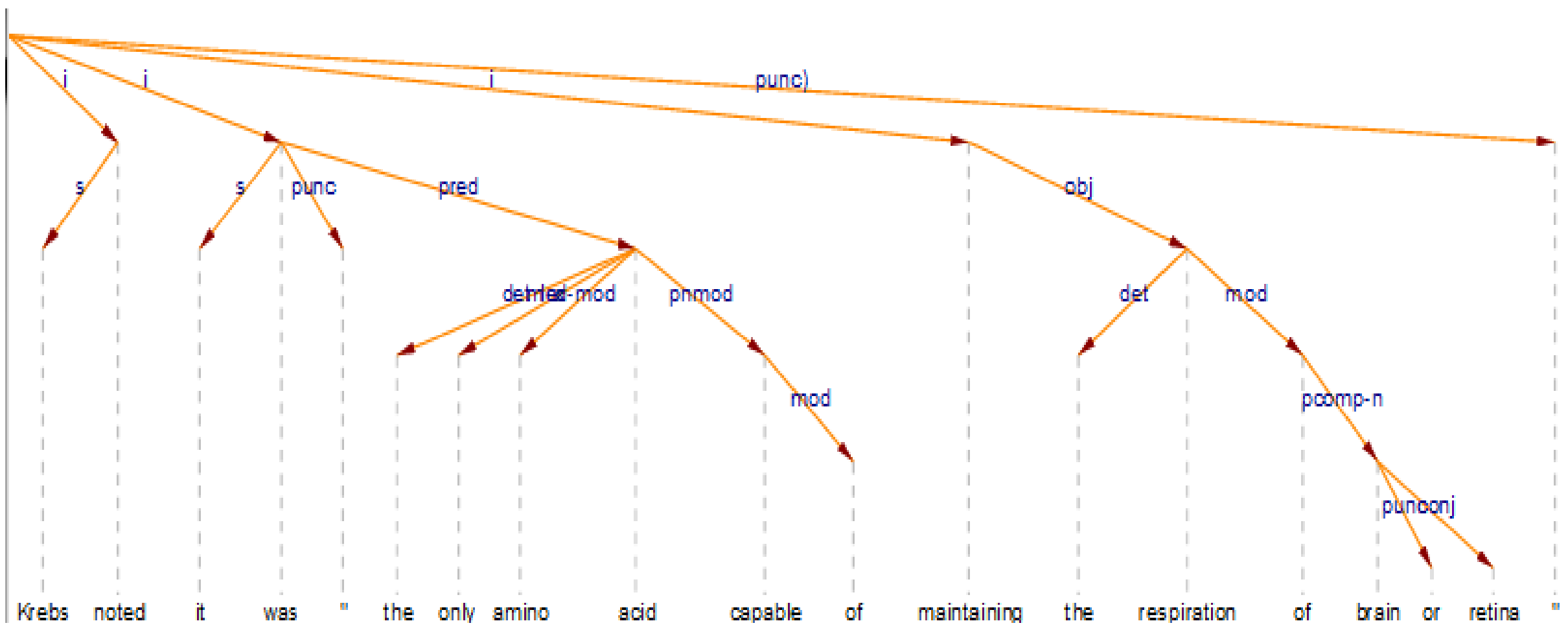

(b)

Figure 6. (a) Stanford and (b) Minipar generate Different Dependency Trees for Quotations

Minipar struggles to generate proper dependency tree and assign wrong dependency relations when it comes across sentences with conjunctions, especially in the form of *NP* and *NP* of *NP*. Stanford Parser performs better than Minipar in such cases. Fig. 7 shows how the parsers generate different dependency tree when they parse sentence with such form.

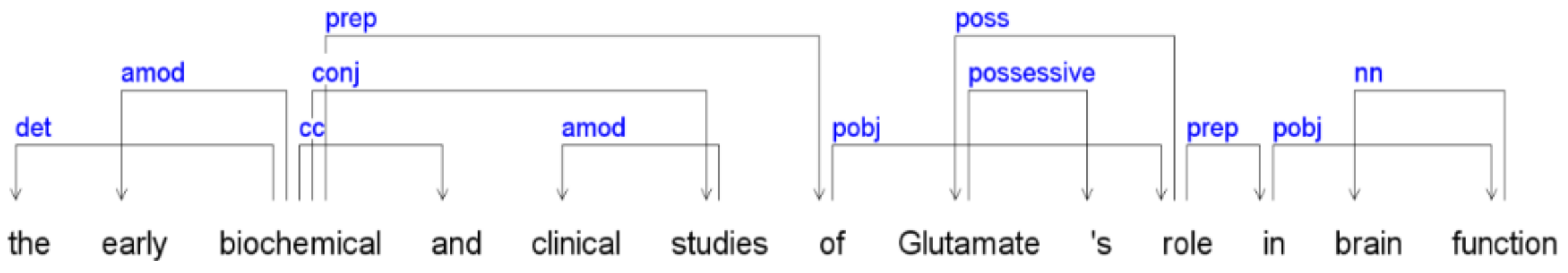

(a)

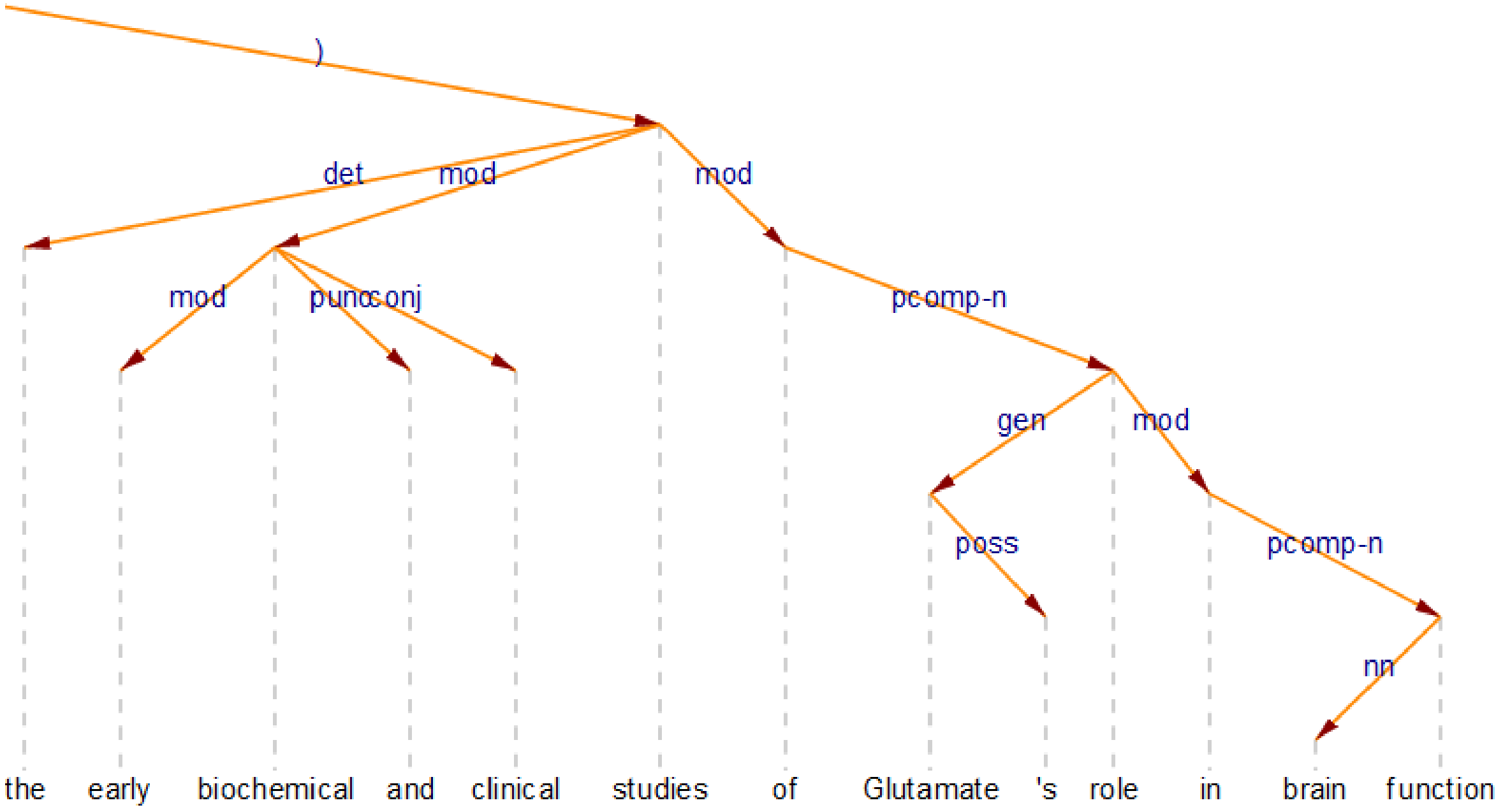

(b)

Figure 7. (a) Stanford and (b) Minipar generate Different Dependency Trees for Sentences with Form *NP* and *NP* of *NP*

Moreover, the parsers differ in producing dependency trees and assigning proper dependencies for the sentences containing WH-clauses as a complement of a preposition. In Fig. 8, we see that the parser outputs begin to differ as soon as they find WH-clause as a complement of a preposition.

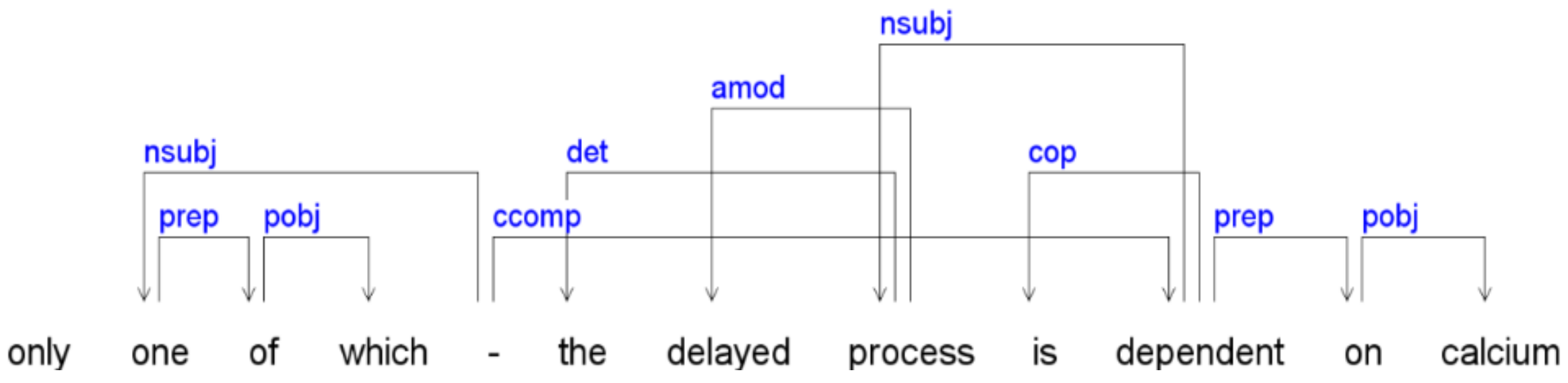

(a)

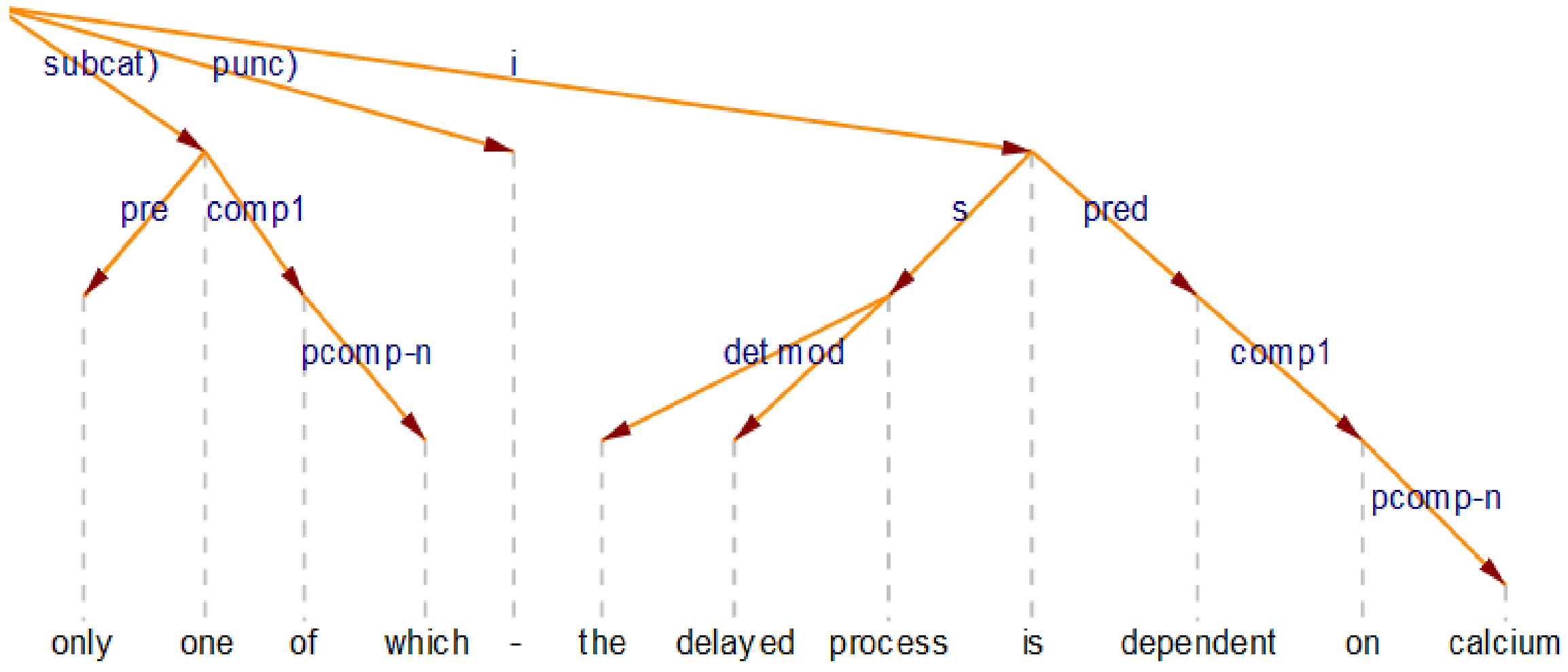

(b)

Figure 8. (a) Stanford and (b) Minipar generate Different Dependency Trees for Sentences with WH-clause as a Complement of Preposition

Besides the difference in working principles of the parsers, the POS tagging of Genia POS tagger is sometimes responsible for inappropriate parsing as well. For example, Genia POS tagger tags the sentence *The thermocouple probe was inserted in the brain striatum* as follows-

*The/DT thermocouple/JJ probe/NN was/VBD inserted/VBN in/IN the/DT brain/NN striatum/NN*

Many would argue, though, to assign *thermocouple* an *NN* tag. Improper tagging leads the parsers to select different heads.

Dependency parsers parse the text and instead of making a clause-structure, develop a Direct Acyclic Graph or DAG. However, Stanford Parser modifies this and is able to produce a DAG with cycles. Minipar does not produce such cyclic DAG. When two parsers in evaluation differ on this aspect of creating cyclic DAG for particular cases, their performance differs as well. For example, Stanford Parser produces a cyclic DAG in Fig. 9. The cycle is manifested for the relations *rcmod (patients, treated)* and *nsubjpass (treated, patients).*

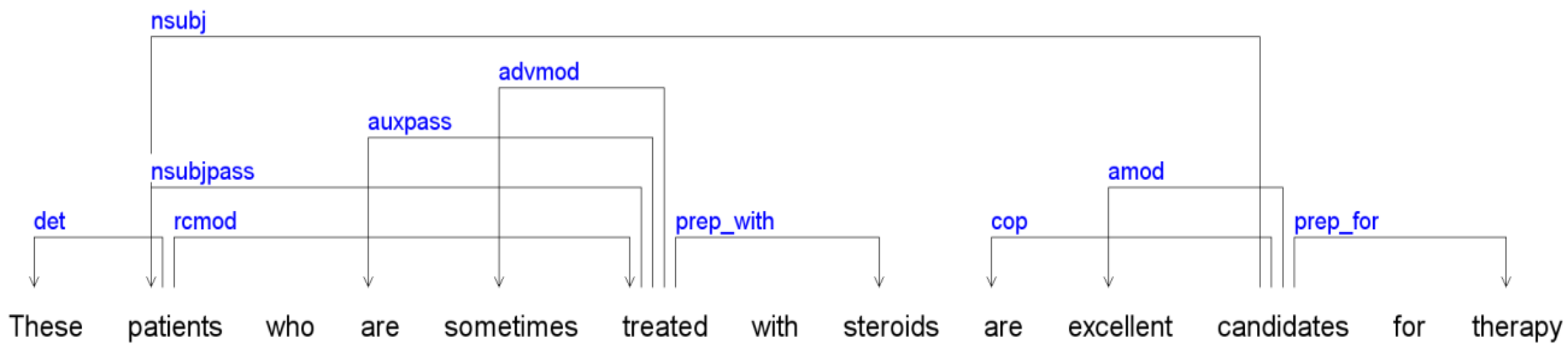

Figure 9. Stanford Parser generates Cyclic DAG

Another reason for the degraded performance of Minipar to parse biomedical texts is our strict evaluation technique. The set of connected concepts was generated without any morphological analysis (e.g., stemming). So, *neuron* and *neurons* are not treated as same word. Though Stanford Parser treats it in the same way, Minipar takes the stem of every word. Therefore, if Stanford Parser selects *neurons* as head of a dependency, Minipar selects *neuron* as its head. In this experiment, strict evaluation is considered- if the heads selected by the parsers did not match fully, it was not counted as an attachment.

Last but not the least, previous research on dependency parser evaluation reported that dependency parsers perform best at the domain of their training corpora [7]. Neither Stanford nor Minipar was trained with a biomedical corpus. This definitely decreases their performances in this domain.

## IV. Conclusions

In this paper, a quantitative and a qualitative analysis on the performance of dependency parsers, namely Stanford and Minipar, to parse biomedical texts have been reported. The experiments showed that the parsers are less successful to find out dependencies between biomedical concepts that are already proved to be connected. The reason for this low success of the parsers is that being statistical parsers, both of them, they cannot find out dependency between words that maintain a decent syntactic distance. The parsing ability for biomedical texts by Minipar was also measured taking the Stanford's as a gold standard in terms of attachment score. The precision, recall and F-Score of the attachment score of Minipar suggest that the parser is not yet ready to parse biomedical texts. However, it is also firmly believed that Minipar's performance on this domain will increase if it is trained with biomedical corpora.

# Appendix

## Source Codes of the Project

1. Stanford Parser Interface

```
/***********************************************************************************************

 * This class takes two concepts from the user and fetches the sentences from the paper that contain both of the terms

 * Then, the sentences are fed to Stanford Parser to get the typed dependencies

 * @author Rushdi Shams

 * @date 06/04/2011

 * @methods:

 * 1. main(): Stanford Parser initialization and dependency extraction

 * Calls FileReader(), ConceptMiner() and WriteFile()

 * 2. FileReader(): Reads the file that has sentences containing both of the concepts

 * 3. WriteFile(): Writes the dependencies into a file

 * 4. ConceptMiner(): Takes the paper and extracts sentences

 * Calls stripGarbage()

 * 5. stripGarbage(): Strips off the garbage character from the sentences extracted from the paper

 * 6. ExtractSentence(): Finds sentences that contain both of the concepts provided by the user

***********************************************************************************************/

import java.io.BufferedReader;

import java.io.BufferedWriter;

import java.io.DataInputStream;

import java.io.File;

import java.io.FileInputStream;

import java.io.FileNotFoundException;

import java.io.FileReader;

import java.io.FileWriter;

import java.io.IOException;

import java.io.InputStreamReader;

import java.text.BreakIterator;

import java.util.*;

import java.util.regex.Matcher;

import java.util.regex.Pattern;

import edu.stanford.nlp.trees.*;

import edu.stanford.nlp.parser.lexparser.LexicalizedParser;
```

```
class ParserDemo {


       public static void main(String[] args) throws IOException {

         LexicalizedParser lp = new LexicalizedParser("H:\\UWO\\PhD\\Stanford Parser\\stanford-parser-2010-08-20\\englishPCFG.ser.gz");

         lp.setOptionFlags(new String[]{"-maxLength", "500", "-retainTmpSubcategories"});

         FileInputStream fstream= new FileInputStream("H:/UWO/Courses/Computational-Linguistics/Dataset/01-Ischemia-Glutamate/file-names.txt");// input file

         DataInputStream in = new DataInputStream(fstream);

         BufferedReader FileNameReader = new BufferedReader(new InputStreamReader(in));

         String FileName="";

         while((FileName = FileNameReader.readLine())!=null){

         try {

                     ConceptMiner(FileName);

               } catch (IOException e) {


                     e.printStackTrace();

               }

         String FileContents = FileReader(FileName);

         String out = "";

         BreakIterator si = BreakIterator.getSentenceInstance();

         si.setText(FileContents);

         int index = 0;

         while (si.next()!=BreakIterator.DONE){

                String test = FileContents.substring(index, si.current());

                String[] sent = test.split(" ");

                Tree parse = (Tree) lp.apply(Arrays.asList(sent));

                TreebankLanguagePack tlp = new PennTreebankLanguagePack();

                GrammaticalStructureFactory gsf = tlp.grammaticalStructureFactory();

                GrammaticalStructure gs = gsf.newGrammaticalStructure(parse);

                Collection tdl = gs.typedDependencies();

                out += EnglishGrammaticalStructure.dependenciesToString(gs, tdl, parse, false, true);

                out +="\n";

                index = si.current();

         }//while (si.next()!=BreakIterator.DONE)

         WriteFile(out, FileName);
```

```
        }//while((FileName = FileNameReader.readLine())!=null)

   }//public static void main(String[] args)



  public static String FileReader(String FileName){

        //File containing all text of a paper

        File file = new File("H:\\UWO\\Courses\\Computational-Linguistics\\Dataset\\01-Ischemia-Glutamate\\"+FileName+".txt");

      StringBuffer contents = new StringBuffer();

      BufferedReader reader = null;

      try{

          reader = new BufferedReader(new FileReader(file));

          String text = null;

          // repeat until all lines is read

          while ((text = reader.readLine()) != null){

              contents.append(text).append(System.getProperty("line.separator"));

          }//while ((text = reader.readLine()) != null)

      }//try

      catch (FileNotFoundException e){

          e.printStackTrace();

      }//catch (FileNotFoundException e)

      catch (IOException e){

          e.printStackTrace();

      }//catch (IOException e)

      finally{

        try{

              if (reader != null){

                  reader.close();

              }//if (reader != null)

          }//try

        catch (IOException e){

              e.printStackTrace();

          }//catch (IOException e)

      }//finally

      String text=contents.toString();

      text = text.trim();

      return text;

        }//public static String FileReader()
```

```
  public static void WriteFile(String out, String FileName){

        try {

             FileWriter fstream = new FileWriter("H:\\UWO\\Courses\\Computational-Linguistics\\Dataset\\01-Ischemia-Glutamate\\"+FileName+"-out.txt");

             fstream.write(out);

             fstream.close();

        }//    try

        catch (IOException e) {

             e.printStackTrace();

       }//catch (IOException e)

  }//public static void WriteFile(String out)


  public static void ConceptMiner(String FileName) throws IOException{

        File file = new File("H:\\UWO\\PhD\\PubMed pairs\\Ischemia-Glutamate\\Papers\\5.txt");

      StringBuffer contents = new StringBuffer();

      BufferedReader reader = null;

      try{

          reader = new BufferedReader(new FileReader(file));

          String text = null;

          // repeat until all lines is read

          while ((text = reader.readLine()) != null){

              contents.append(text).append(System.getProperty("line.separator"));

          }//while ((text = reader.readLine()) != null)

      }//try

      catch (FileNotFoundException e){

          e.printStackTrace();

      }//catch (FileNotFoundException e)

      catch (IOException e){

          e.printStackTrace();

      }//catch (IOException e)

      finally{

       try{

             if (reader != null){

                 reader.close();

             }//if (reader != null)

         }//try
```

```java
        catch (IOException e){

              e.printStackTrace();

           }//catch (IOException e)

       }//finally

       System.out.println("Text from "+file+" has been extracted...\n");

       /* Taking the file contents to a variable named text and sending that to a method to strip
garbage characters*/

       String text=contents.toString();

       String strippedtext=stripGarbage(text);

       if(strippedtext!=null){

       System.out.println("Garbage characters have been stripped off...\n");

       System.out.println(strippedtext);

       }//if(strippedtext!=null)

       //sending stripped text to extractsentence method of sentenceExtract class

       System.out.println("Now, applying a keyword based search...\n");

       ExtractSentence(strippedtext, FileName);

   }//public static void ConceptMiner()

   public static String stripGarbage(String s) {

        String valid =" abcdefghijklmnopqrstuvwxyzABCDEFGHIJKLMNOPQRSTUVWXYZ.";

        String strippedtext = "";

        for ( int i = 0; i < s.length(); i++ ) {

               if ( valid.indexOf(s.charAt(i)) >= 0 ){

                      strippedtext += s.charAt(i);

               }//if ( valid.indexOf(s.charAt(i)) >= 0 )

           }//for ( int i = 0; i < s.length(); i++ )

        return strippedtext;

        }//public static String stripGarbage(String s)

   public static void ExtractSentence(String text, String FileName) throws IOException{

        String sentence=null;

        int index = 0,bothcount=0;

        Scanner input = new Scanner( System.in );

        System.out.print( "Concept one (in small letter): " );

        String keywordone=input.next();

        System.out.print( "Concept two (in small letter): " );

        String keywordtwo=input.next();

        Pattern p = Pattern.compile(keywordone);

        Pattern q = Pattern.compile(keywordtwo);
```

```
        BreakIterator bi = BreakIterator.getSentenceInstance();

        bi.setText(text);

        FileWriter      fstreamboth      =      new      FileWriter("H:\\UWO\\Courses\\Computational-
Linguistics\\Dataset\\01-Ischemia-Glutamate\\"+FileName+".txt");

        BufferedWriter outputboth = new BufferedWriter(fstreamboth);

        while (bi.next() != BreakIterator.DONE) {

                sentence = text.substring(index, bi.current());

                sentence = sentence.trim();

                sentence = sentence.toLowerCase();

                Matcher matcherone = p.matcher(sentence);

                Matcher matchertwo = q.matcher(sentence);

                if (matcherone.find()) {

                        if(matchertwo.find()){

                                outputboth.write(sentence);

                                outputboth.write("\n\n");

                                bothcount++;

                                sentence=null;

                        }//if(matchertwo.find())


                }//if (matcherone.find())

                else if (matchertwo.find()) {

                        if(matcherone.find()){

                                outputboth.write(sentence);

                                outputboth.write("\n\n");

                                bothcount++;

                                sentence=null;

                        }//if(matcherone.find())

                }//else if (matchertwo.find())

                index = bi.current();

        }  //  while (bi.next() != BreakIterator.DONE)

        outputboth.close();

        System.out.println(bothcount+" sentences with both of the terms");

        }//public static void extractSentence(String text) throws IOException

}// class ParserDemo
```

2. Minipar Interface

```
import java.io.BufferedReader;

import java.io.File;

import java.io.FileNotFoundException;

import java.io.FileReader;

import java.io.FileWriter;

import java.io.IOException;

import java.text.BreakIterator;


/**************************************************************************
 * This is the JNI proxy class for accessing the Minipar library.
 * @author Rushdi Shams
 * @date 06/04/2011
 *************************************************************************/


public class MiniParProject {

	// These three methods are implemented in MiniparCpp/MiniparProxy.cpp

	public native boolean InitMinipar(String datapath);

	public native String Parse(String line);

	private MiniParProject(){} // not allowed to construct without libpath

	// libpath is the full path to libMiniparProxy.so

	public  MiniParProject( String libpath ) {

		System.load(libpath);

	}

	// Method for testing this class

	public static void main(String[] args) throws IOException {

		String libpath = "C:\\Rushdi\\MiniparProxy\\MiniparCpp\\MiniparProxy.dll";

		MiniparProxy mp = new MiniparProxy(libpath);


		System.out.println(String.valueOf(mp.InitMinipar("C:\\Rushdi\\MiniparProxy\\minipar
		\\data")));

		String FileName="10-cl-gradient";

		FileWriter fw = new FileWriter("C:\\Rushdi\\"+FileName+"-minipar.txt");

		int cnt = 1;

		int index = 0;

		String out = "";

		String FileContents = FileReader(FileName);

		BreakIterator si = BreakIterator.getSentenceInstance();
```

```java
        si.setText(FileContents);
        while (si.next()!=BreakIterator.DONE){
               String s = FileContents.substring(index,si.current());
               String re = mp.Parse( s );
               System.out.println( re );
               out += re;
               out += "\n";
        }
        index = si.next();
        }
        fw.write(out);
        fw.close();
}
  public static String FileReader(String FileName){
         //File containing all text of a paper
         File file = new File("C:\\Rushdi\\Dataset\\04-Epilepsy-GABA\\"+FileName+".txt");
      StringBuffer contents = new StringBuffer();
      BufferedReader reader = null;
      try{
          reader = new BufferedReader(new FileReader(file));
          String text = null;
          while ((text = reader.readLine()) != null){
              contents.append(text).append(System.getProperty("line.separator"));
          }//while ((text = reader.readLine()) != null)
      }//try
      catch (FileNotFoundException e){
          e.printStackTrace();
      }//catch (FileNotFoundException e)
      catch (IOException e){
          e.printStackTrace();
      }//catch (IOException e)
      finally{
        try{
              if (reader != null){
                  reader.close();
              }//if (reader != null)
          }//try
```

```
               catch (IOException e){

                      e.printStackTrace();

                  }//catch (IOException e)

              }//finally

              String text=contents.toString();

              text = text.trim();

              return text;

               }//public static String FileReader()

}
```

## 3. Interface to evaluate Minipar Parser

```
/*************************************************************************************************
 * This class evaluates minipar as a dependency parser for biomedical domain by maintaining
 * stanford parser's output as gold standard
 * To evaluate, the number of heads produced by Stanford was compared against the number of heads
 * produced by minipar. Evaluation metric is precision and recall.
 * @author Rushdi Shams
 * @date 16/04/2011
 * @methods:
 * 1. main(): Takes the stanford and minipar output file from the user
 * Calls FileReader (), MiniparHeadDetermine (), StanHeadDetermine () and EvaluateParser ()
 * 2. FileReader(): Reads the files: reads stanford output file and minipar output file
 * 3. MiniparHeadDetermine(): Determines Heads from minipar output and takes them to a string
 * 4. StanHeadDetermine(): Determines Heads from Stanford output and takes them to a string
 * 5. EvaluateParser (): Determines precision and recall of minipar parser
*************************************************************************************************/

import java.io.BufferedReader;

import java.io.File;

import java.io.FileNotFoundException;

import java.io.FileReader;

import java.io.FileWriter;

import java.io.IOException;

import java.util.Collection;

import java.util.Iterator;

import java.util.StringTokenizer;

import java.util.TreeMap;
```

```
public class ParserEvaluation {

	public static void main(String[] args){

		String StanoutFileName = "10-cl-gradient-out";

		String MiniparoutFileName = "10-cl-gradient-minipar";

		String StanoutContent = FileReader (StanoutFileName);

		String MiniparoutContent = FileReader (MiniparoutFileName);

		String StanHead = StanHeadDetermine (StanoutContent);

		System.out.println(StanHead);

		String MiniparHead = MiniparHeadDetermine(MiniparoutContent);

		System.out.println(MiniparHead);

		EvaluateParser (StanHead, MiniparHead);

	}// public static void main(String[] args)

	public static void EvaluateParser (String StanHead, String MiniparHead){

	// Treemap for Stanford Parser Heads and their frequency

		final TreeMap<String, Integer> StanHeadMap = new TreeMap<String, Integer>();

		// Iterate through each word of the current line...Delimit words based on whitespace

		final StringTokenizer ST_Stanford = new StringTokenizer(StanHead, " ");

		// for all heads

		while (ST_Stanford.hasMoreTokens()) {

			final String currentWord = ST_Stanford.nextToken();

			Integer frequency = StanHeadMap.get(currentWord);

			// Add the word if it doesn't already exist, otherwise increment the frequency counter.

			if (frequency == null) {

				frequency = 0;

			} //if (frequency == null)

			StanHeadMap.put(currentWord, frequency + 1); // inserting the head and its frequency

		}//while (ST_Stanford.hasMoreTokens())

		System.out.println(StanHeadMap);

		// Treemap for Minipar Parser Heads and their frequency

		final TreeMap<String, Integer> MiniparHeadMap = new TreeMap<String, Integer>();

		//Iterate through each word of the current line...Delimit words based on whitespace

		final StringTokenizer ST_Minipar = new StringTokenizer(MiniparHead, " ");

		//for all heads

		while (ST_Minipar.hasMoreTokens()) {

			final String currentWord = ST_Minipar.nextToken();
```

```
                    Integer frequency = MiniparHeadMap.get(currentWord);
                    //     Add the word if it doesn't already exist, otherwise increment the frequency counter.
                    if (frequency == null) {
                        frequency = 0;
                    } //if (frequency == null)
                    MiniparHeadMap.put(currentWord, frequency + 1);
                }//while (ST_Minipar.hasMoreTokens())
                System.out.println(MiniparHeadMap);
                // setting iterators for minipar keys and values and stanford keys and values
            Iterator st_v_itr = StanHeadMap.values().iterator();
            Iterator m_v_itr = MiniparHeadMap.values().iterator();
            Iterator st_k_itr = StanHeadMap.keySet().iterator();
            Iterator m_k_itr = MiniparHeadMap.keySet().iterator();
            int fp = 0, tp = 0, fn = 0; // initializing positives and negatives
            // for every stanford head
            while (st_k_itr.hasNext()){
                String ST_Key = (String) st_k_itr.next(); // we get the stanford head
                int ST_Value = (Integer) st_v_itr.next(); // we get the head's frequency
                if (MiniparHeadMap.containsKey(ST_Key)){ // if minipar tree also contains the head
                    int  M_Value  =  MiniparHeadMap.get(ST_Key);  //  get  the  frequency  of  the minipar head
                    int diff = ST_Value - M_Value; // the stanford head frequency and minipar head frequency are subtracted
                    if (diff == 0){ // if no difference
                        tp = tp + ST_Value; // stanford parser's head frequency is the tp
                        fn = fn + 0; // the rest does not matter
                        fp = fp + 0;// the rest does not matter
                    }//if (diff == 0)
                    else if (diff <0){ //if minipar head has more frequency
                        tp = tp + ST_Value; // true positive is the stanford head frequency
                        fp  =  fp  +  Math.abs(diff);  //  the  extra  (differenc)  is  the  false positive
                        fn = fn + 0; // the rest does not matter
                    }//else if (diff <0)
                    else if (diff > 0){ // if stanford produces more head frequency
                        tp = tp + M_Value; // then, minipar head frequency is true positives
                        fn = fn + diff; // false negative is the difference between them
                        fp = fp + 0; // the rest does not matter
```

```
                    }//else if (diff > 0)

                }//if (MiniparHeadMap.containsKey(ST_Key))

                else

                    fn = fn + ST_Value; // if minipar misses the head, the stanford head frequency is the false negative

            }//while (st_k_itr.hasNext())

            // we have to do the same for the reverse

            st_v_itr = StanHeadMap.values().iterator();

            m_v_itr = MiniparHeadMap.values().iterator();

            st_k_itr = StanHeadMap.keySet().iterator();

            m_k_itr = MiniparHeadMap.keySet().iterator();

            while (m_k_itr.hasNext()){

                String M_Key = (String) m_k_itr.next();

                int M_Value = (Integer) m_v_itr.next();

                if (!StanHeadMap.containsKey(M_Key)){ // if minipar produces a head, that is not in stanford head list

                    fp = fp + M_Value;// all of them are false positives

                    tp = tp + 0;// there is nothing to do with the rest

                    fn = fn + 0;

                }//if (!StanHeadMap.containsKey(M_Key))

            }//while (m_k_itr.hasNext())

            System.out.println ("TP: "+tp+" FP: "+fp+" FN: "+fn);

            double param1 = tp + fp;

            double param2 = tp + fn;

            System.out.println(param1 + " " + param2);

            double precision = 0.0, recall = 0.0;

            precision = (tp / param1) * 100;

            recall = (tp / param2) * 100;

            System.out.println ("Precision: " + precision + " Recall: " + recall);

            try {

                    FileWriter fw = new FileWriter ("H:\\UWO\\Courses\\Computational-Linguistics\\Dataset\\04-precision-recall.txt", true);

                    fw.write("Precision " + precision + " Recall " + recall + "\n");

                    fw.close();

                } catch (IOException e) {

                    e.printStackTrace();

                }

        }//public static void EvaluateParser (String StanHead, String MiniparHead)
```

```
public static String StanHeadDetermine (String StanoutContent){
        String good =" abcdefghijklmnopqrstuvwxyzABCDEFGHIJKLMNOPQRSTUVWXYZ(";
        String result = "";
    for ( int i = 0; i < StanoutContent.length(); i++ ) {
        if ( good.indexOf(StanoutContent.charAt(i)) >= 0 )
           result += StanoutContent.charAt(i);
    }
    System.out.println(result);
    String [] temp = result.split(" ");
    String StanHead = "";
    for (int i = 0;i < temp.length-1; i++){
        String temp2 = temp[i].toString().replace('(', ' ');
        System.out.println(temp2);
        String [] temp3 = temp2.split(" ");
        if (temp3[1] != null){
                StanHead += temp3[1]+" ";
        }
        else continue;
    }
return StanHead;
}//public static String StanHeadDetermine (String StanoutContent)
public static String MiniparHeadDetermine (String MiniparoutContent){
        String in = MiniparoutContent.replaceAll("\t", " ");
        String good =" abcdefghijklmnopqrstuvwxyzABCDEFGHIJKLMNOPQRSTUVWXYZ._";
        String result = "";
    for ( int i = 0; i < in.length(); i++ ) {
        if ( good.indexOf(in.charAt(i)) >= 0 )
           result += in.charAt(i);
    }//for ( int i = 0; i < in.length(); i++ )
    String MiniparHead = "";
    String [] temp = result.split(" ");
    for (int j = 0; j<temp.length;j++)
        if (temp [j].contentEquals("gov")){
                MiniparHead += temp[j+1]+" ";
    }//for (int j = 0; j<temp.length;j++)
   return MiniparHead;
}//public static String MiniparHeadDetermine (String MiniparContent)
```

```
        public static String FileReader(String FileName){

                //File containing all text of a paper

                File  file  =  new  File("H:\\UWO\\Courses\\Computational-Linguistics\\Dataset\\04-Epilepsy-GABA\\"+FileName+".txt");

            StringBuffer contents = new StringBuffer();

            BufferedReader reader = null;

           try{

                   reader = new BufferedReader(new FileReader(file));

                    String text = null;

               // repeat until all lines is read

               while ((text = reader.readLine()) != null){

                   contents.append(text).append(System.getProperty("line.separator"));

               }//while ((text = reader.readLine()) != null)

           }//try

           catch (FileNotFoundException e){

                   e.printStackTrace();

           }//catch (FileNotFoundException e)

           catch (IOException e){

                   e.printStackTrace();

           }//catch (IOException e)

           finally{

                   try{

                           if (reader != null){

                                   reader.close();

                    }//if (reader != null)

               }//try

                   catch (IOException e){

                           e.printStackTrace();

               }//catch (IOException e)

           }//finally

           String text=contents.toString();

           text = text.trim();

           return text;

        }//public static String FileReader()

}//public class ParserEvaluation
```